\pdfoutput=1

\documentclass[11pt]{article}

\usepackage[final]{acl}
\usepackage{booktabs}
\usepackage{times}
\usepackage{latexsym}
\usepackage{enumitem}
\usepackage{float}
\usepackage{amsfonts}
\usepackage{amsmath}
\usepackage[T1]{fontenc}

\usepackage[utf8]{inputenc}

\usepackage{microtype}

\usepackage{inconsolata}

\usepackage{graphicx}
\usepackage{tikz}
\pdfoutput=1
\usepackage[edges]{forest}
\usepackage{xcolor}
\usetikzlibrary{shapes.geometric, arrows}
\definecolor{hidden-black}{RGB}{20,68,106}
\definecolor{purple}{RGB}{144,153,196}
\definecolor{yellow}{RGB}{255,228,123}
\newcommand{\eg}{\textit{e.g.,}}
\usepackage{booktabs}  
\usepackage{makecell}  
\usepackage{pifont}
\newcommand{\cmark}{\ding{51}}%
\newcommand{\xmark}{\ding{55}}%
\title{Large Vision-Language Model Alignment and Misalignment:\\ A Survey Through the Lens of Explainability
}

\author{Dong Shu\textsuperscript{1}, Haiyan Zhao\textsuperscript{2}, Jingyu Hu\textsuperscript{3}, Weiru Liu\textsuperscript{3}, 
Ali Payani\textsuperscript{4},\\
\textbf{Lu Cheng\textsuperscript{5}}, \textbf{Mengnan Du\textsuperscript{2}\thanks{Corresponding author.}}\\
\textsuperscript{1}Northwestern University,
\textsuperscript{2}New Jersey Institute of Technology,\\
\textsuperscript{3}University of Bristol,
\textsuperscript{4}Cisco Research,\textsuperscript{5}{University of Illinois Chicago}\\
\small\texttt{dongshu2024@u.northwestern.edu},\, \small\texttt{mengnan.du@njit.edu}
}

\begin{document}
\maketitle
\begin{abstract}
Large Vision-Language Models (LVLMs) have demonstrated remarkable capabilities in processing both visual and textual information. However, the critical challenge of alignment between visual and textual representations is not fully understood. This survey presents a comprehensive examination of alignment and misalignment in LVLMs through an explainability lens.
We first examine the fundamentals of alignment, exploring its representational and behavioral aspects, training methodologies, and theoretical foundations. We then analyze misalignment phenomena across three semantic levels: object, attribute, and relational misalignment. Our investigation reveals that misalignment emerges from challenges at multiple levels: the data level, the model level, and the inference level.
We provide a comprehensive review of existing mitigation strategies, categorizing them into parameter-frozen and parameter-tuning approaches. Finally, we outline promising future research directions, emphasizing the need for standardized evaluation protocols and in-depth explainability studies. 
\end{abstract}

\section{Introduction}

Recent Large Vision-Language Models (LVLMs) have achieved significant progress in multimodal understanding. Models such as GPT-4V \cite{gpt4v}, Gemini \cite{team2023gemini}, LLaVA \cite{liu2024visual}, Claude-3.5-Sonnet~\cite{anthropic2024claude}, Qwen2-VL~\cite{wang2024qwen2}, and LLaMa 3.2 \cite{dubey2024llama} demonstrate remarkable capabilities in tasks like image captioning and visual question answering, not only processing visual and textual information independently but also reasoning across these modalities. These advances are built upon two key foundations: large language models (LLMs) and vision encoders. LLMs such as GPT-3.5 \cite{brown2020language}, LLaMA \cite{touvron2023llama}, LLaMA 2 \cite{touvron2023llama2}, and Qwen \cite{bai2023qwen} transformed natural language processing, while visual understanding models like Vision Transformer (ViT)~\cite{dosovitskiy2021image} and vision-language models like CLIP~\cite{radford2021learning} have advanced the ability to process visual information and create aligned visual-textual representations respectively.

The key challenge in developing effective LVLMs lies in achieving proper alignment between visual and textual representations \cite{liu2024survey}. The predominant approach involves using representation alignment techniques, where visual features from an image encoder and textual representations from an LLM are mapped into a shared embedding space, typically matching the LLM's embedding dimensions~\cite{jia2021scaling, yang2022vision, shu2024exploring}. Once both modalities are mapped into this shared space, alignment can be achieved through various training objectives and architectural designs that encourage the model to understand and reason about cross-modal relationships. This method has gained popularity due to its straightforward approach and generalizability across different model architectures.

However, the current understanding of alignment mechanisms remains limited. A critical challenge lies in misalignment phenomena, which manifest in various forms. For instance, when shown an image of a green apple, the model might fail to recognize the apple altogether (object misalignment), incorrectly describe it as red (attribute misalignment), or generate incorrect relationships like ``the apple is floating in the air'' when it's sitting on a table (relational misalignment). These misalignments lead to reliability issues \cite{zhang2024rethinking, zhou2024aligning,zhao2024survey}, where models generate textual outputs that are inconsistent with the visual input. Understanding and addressing these misalignment issues is crucial for developing more reliable and trustworthy LVLMs, as they directly impact the models' ability to generate accurate and consistent multimodal outputs.

\subsection{Contribution and Uniqueness}
\noindent\textbf{Our Contributions.}
In this survey, we present a structured framework for understanding and addressing alignment challenges in LVLMs from an explainability perspective. Our major contributions are listed as follows:
(1) We examine the fundamentals of alignment, covering its representational and behavioral aspects, training procedures, and theoretical foundations (Section~\ref{sec:alignment_of_lvlms}). 
(2) We analyze misalignment at both the representation and behavior levels, and categorize behavioral phenomena into object, attribute, and relational misalignment (Section~\ref{sec:misalignment_of_lvlms}). 
(3) We identify that misalignment arises from challenges at three key levels: data level, model level, and inference level (Section~\ref{sec:misalignment_of_lvlms}). 
(4) We review existing mitigation strategies with an emphasis on their underlying motivations, trade-offs, and assumptions, and further enhance the explainability analysis of these approaches (Section~\ref{sec:mitigation}).
(5) In the Appendix, we provide extended discussions including existing evaluation methods, real-world examples of LVLM misalignment, and mitigation methods comparative analysis.

\vspace{3pt}
\noindent\textbf{Differences with Existing Surveys.}
While several existing surveys focus primarily on hallucination in LVLMs~\cite{liu2024survey, bai2024hallucination, sahoo2024comprehensive}, our work addresses the broader and more foundational concept of misalignment. We view hallucination as one possible consequence of misalignment. Another key distinction is our focus on explainability as the central lens for understanding alignment. We systematically analyze alignment through the questions of \emph{what} alignment is, \emph{how} it is achieved, and \emph{why} it is possible. This explainability-driven perspective guides our entire analysis, including the design of our mitigation taxonomy and the in-depth discussion of existing mitigation methods.

\section{Alignment of LVLMs}
\label{sec:alignment_of_lvlms}
In this section, we examine alignment in LVLMs across four essential dimensions. First, we define the concept of alignment in LVLMs. Second, we detail the procedural stages through which alignment is achieved in practice. Third, we explore the theoretical foundations that make alignment possible between visual and textual modalities. Finally, we discuss methods for measuring and evaluating alignment in LVLMs in Appendix \ref{appendix:measure_alignment}.

\subsection{What is Alignment?}
In the context of LVLMs, let $\mathcal{X}$ be the image space and $\mathcal{T}$ be the text space. We define the alignment in two fundamental aspects: representational alignment and behavioral alignment.

\begin{itemize}[leftmargin=*]\setlength\itemsep{-0.3em}
\item
\emph{Representational alignment} refers to the degree of correspondence between visual representations $v \in \mathcal{V}$ and textual representations $t \in \mathcal{T}$ within the model's internal embedding space $\mathcal{E}$. When well-aligned, the visual features extracted from an image and the textual embeddings of its corresponding description occupy nearby regions in the shared latent space, exhibiting high semantic similarity $d(v,t)$ where $d$ is a similarity metric. This internal alignment enables the model to establish meaningful connections between visual and textual information at a fundamental level.
\item 
\emph{Behavioral alignment} refers to the model's ability to generate accurate, factual, and consistent textual responses $y \in \mathcal{Y}$ when processing image inputs $x \in \mathcal{X}$. A behaviorally aligned LVLM can reliably answer questions about visual content, provide precise descriptions, and perform reasoning tasks without introducing errors or hallucinations. This external manifestation ensures that the model's outputs faithfully reflect the actual content and relationships present in the images.
\end{itemize}

These two aspects of alignment are inherently connected. Strong representational alignment typically supports better behavioral alignment, as the model can more effectively leverage both visual and textual information to generate reliable outputs. Conversely, poor alignment in either aspect can lead to issues such as mismatched representations, inaccurate responses, or hallucinated content.

\subsection{Why is Alignment Possible?}

Having established what alignment means and how it is implemented in LVLMs, a fundamental question arises: why is such alignment between vision and language modalities possible in the first place? The possibility of alignment between these modalities can be understood from both theoretical and algorithmic perspectives.

\vspace{3pt}
\noindent\textbf{Theoretical Perspective.}\,
From a theoretical standpoint, visual and textual data are different projections of the same underlying reality. As Huh et al. argue in their Platonic Representation Hypothesis~\cite{huh2024platonic}, all modalities are measurements of a real world that generates our observations. When humans create images or write text, they are encoding information about this same reality, though through different measurement processes. Although these modalities appear distinct on the surface, they fundamentally capture overlapping semantic information about the same world state. This shared origin in physical reality, combined with the fact that humans generate both types of data to describe their observations of the world, provides the theoretical foundation for why these modalities can be meaningfully aligned in a common representation space. See Appendix~\ref{appendix:theoretical_perspective} for a formal justification.

\vspace{3pt}
\noindent\textbf{Algorithmic Perspective.}\,
From an algorithmic perspective, although visual encoders and language models are initially trained separately on different modality-specific data, their learned representations inherently capture some similar semantic structures due to their training on human-generated data. Recent research has shown that these inherent similarities exist even before explicit alignment training~\cite{maniparambil2024vision, sharma2024vision, neo2024towards}. This natural compatibility serves as a starting point for more sophisticated alignment. The staged training process described in Section 2.2 then is built upon this inherent compatibility through systematic refinement: first using contrastive learning to organize embeddings in the shared latent space, then employing adapter fine-tuning to bridge between modalities while preserving their specialized capabilities, and finally conducting end-to-end training to enable deep integration across all components. Through this systematic combination of training stages and optimization objectives, the model gradually develops a robust alignment between the two modalities.

\begin{figure*}
    \centering
    \includegraphics[width=1\linewidth]{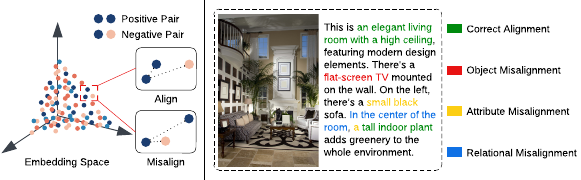}
    \caption{
    Illustration of representation-level and behavior-level alignment and misalignment in LVLMs. The \textbf{left} side shows \textbf{representation-level} phenomena in embedding space, where aligned visual-text pairs cluster together (positive pairs) while misaligned pairs are separated (negative pairs). The \textbf{right} side demonstrates \textbf{behavior-level} alignment and misalignment through a room description example, showing the spectrum from correct alignment (green) to various types of semantic misalignment: object misalignment (red), attribute misalignment (yellow), and relational misalignment (blue). These two levels are inherently connected, as the quality of representation alignment in the embedding space influences the model's ability to generate semantically aligned outputs.
    }
    \label{fig:misalignment}
\end{figure*}

\section{Misalignment of LVLMs}
\label{sec:misalignment_of_lvlms}
After introducing the alignment of LVLMs, we now examine a critical challenge facing these models: their tendency to generate outputs that diverge from the visual input. Despite significant advances in alignment techniques, LVLMs still frequently exhibit misalignment between their visual and textual inputs. In this section, we provide a comprehensive analysis of misalignment phenomena in LVLMs, beginning with a definition and taxonomy of different types of misalignment (see Figure \ref{fig:misalignment}), followed by an examination of their underlying causes.


\subsection{Definition of Misalignment}
Misalignment in LVLMs occurs when the model's output semantically diverges from the visual content it is meant to describe. These discrepancies show in several key phenomena, impacting the overall performance of these models.
In this section, we categorize behavior-level misalignment phenomena in LVLM into three semantic levels $\mathcal{S} = \{s_o, s_a, s_r\}$: \emph{object misalignment} ($s_o$), \emph{attribute misalignment} ($s_a$), and \emph{relation misalignment} ($s_r$) (Figure \ref{fig:misalignment} right). Rather than using the term `hallucination' commonly found in the literature~\cite{liu2024survey}, we adopt the term `misalignment' to better characterize how these discrepancies emerge between visual and language representations. For representation-level misalignment, please refer to Figure \ref{fig:misalignment} left and Appendix \ref{appendix:representation_misalignment}.

\begin{itemize}[leftmargin=*]\setlength\itemsep{-0.3em}
\item \emph{Object Misalignment} ($s_o$): This is one of the most widely recognized forms of misalignment \cite{liu2024survey, wang2023evaluation, li2023evaluating}. It occurs when the model generates descriptions containing objects $O'$ that differ from the actual objects $O$ in the image, where $O' \not\subseteq O$. This represents the most coarse-grained level of misalignment, as it simply refers whether an object exists in the image or not. Due to its coarse-grained nature, object misalignment is relatively straightforward to detect and mitigate. 

\item \emph{Attribute Misalignment} ($s_a$): At a finer level, we identify attribute misalignment \cite{shang2024pixels}. This occurs when for an object $o \in O$, the model correctly identifies the object but generates incorrect attributes $A' \neq A$, where $A$ represents the true attributes of $o$. Attribute misalignment typically involves adjectives or adverbs that describe properties of objects inaccurately. For example, when input an image of a green apple, the model might incorrectly describe the color of an apple as `red' instead of `green'. 

\item \emph{Relation Misalignment} ($s_r$): This category involves the generation of incorrect or non-existent relationships $R'$ between objects in an image~\cite{wu2024evaluating}, where $R'$ differs from the true relationships $R$. This misalignment manifests in two primary ways: spatial relationship errors and action relationship errors. In spatial relationships, the model might incorrectly describe the relative positions of objects, such as saying `next to' when the correct relation is `on top of', or `inside' when objects are merely `near' each other. In action relationships, the model might generate semantically impossible interactions between objects, such as `he is walking a car' instead of `he is driving a car', or `the cat is reading a book' instead of `the cat is sitting on a book'.  
\end{itemize}


To better illustrate real-life examples of misalignment, we have selected four prominent LVLMs and provided their generated responses in Appendix \ref{appendix:example}.

\subsection{Reasons of Misalignment}
Having identified the three semantic levels of misalignment phenomena, we now analyze their root causes across three fundamental levels: Dataset, Model, and Inference. The Dataset level examines how training data characteristics influence misalignment during learning. The Model level investigates how architectural decisions and training procedures affect alignment between modalities. The Inference level explores how the generation process can introduce misalignment even with well-aligned underlying representations.

\subsubsection{Dataset Level}\label{sec:data-level-issue}
Data quality and distribution patterns play crucial roles in contributing to misalignment between visual and language representations in LVLMs. Several key dataset factors can impede the model's ability to form accurate associations between visual inputs and textual descriptions, affecting both training effectiveness and inference performance. We list one factor below, for additional factors please refer to Appendix \ref{appendix:more_reason}.

\begin{itemize}[leftmargin=*]\setlength\itemsep{-0.3em}
\item 
\emph{Data imperfections}:
This includes blurry images, vague or inaccurate captions, and mismatched image-caption pairs, which introduce significant challenges during training~\cite{ouali2025clip, shi2024assessment}. These quality issues manifest in various forms: images may suffer from poor resolution, inappropriate cropping, or visual artifacts; captions might contain grammatical errors, ambiguous descriptions, or factually incorrect information; and in some cases, the captions may describe content entirely unrelated to their paired images. These low-quality data points can distort the model's ability to form precise mappings between modalities, leading to outputs that fail to accurately reflect the input image and potentially establishing incorrect associations that persist through the training process.
\end{itemize}

\subsubsection{Model Level}
Beyond data-level issues, the architectural design and training methodology of LVLMs significantly influence model alignment. 
\begin{itemize}[leftmargin=*]\setlength\itemsep{-0.3em}

\item
\emph{Ability Gap}:
This independent pretraining process also creates an ability gap between the visual encoder and the LLM \cite{li2024enhancing}, where the LLM often demonstrates significantly greater capability than the visual encoder. Consequently, the LVLM tends to rely excessively on the LLM for predictions, resulting in imbalanced attention between visual and textual information \cite{chen2025image, min2024mitigating, woo2024don}. 

\item
\emph{Pretrain-finetuning Knowledge Gap}:
After integrating the visual encoder and LLM into a unified LVLM, fine-tuning is typically performed to further enhance alignment and adapt the model to specific downstream tasks. However, this fine-tuning phase can introduce a pretraining-finetuning knowledge gap or conflict, where the general knowledge acquired during pretraining may clash with the specific requirements of the fine-tuning task \cite{zhou2024aligning}. Such conflicts can lead to knowledge forgetting, where the LVLM loses previously learned information while adapting to the new task \cite{zhou2023learning, huang2024learn}. Although knowledge forgetting might appear insignificant, it can have cascading effects. Each unit of knowledge in the model’s embedding space is interconnected with lots of semantic relationships. Forgetting even a single piece of knowledge can disrupt these relational connections, undermining the integrity of the embedding space. This disruption causes a broader misalignment within the LVLM.

\item
\emph{Knowledge Conflict}:
A significant challenge arises from knowledge conflicts between the visual knowledge of image and parametric knowledge of LLM. These conflicts emerge when the visual encoder's direct perception of image content contradicts the prior knowledge embedded in the LLM's parameters during pre-training~\cite{zhu2024unraveling, ghosh2024visual}. For example, when an image contains a green tomato, the visual encoder accurately detects its color, but the LLM may resist this information since it has been predominantly trained on texts describing ripe, red tomatoes. This misalignment between observed visual evidence and learned textual priors can manifest in various ways: the model might incorrectly describe the tomato as red despite clear visual evidence, generate hesitant or self-contradicting descriptions, or attempt to rationalize the discrepancy by making unwarranted assumptions about the tomato's ripeness stage. 
\end{itemize}

\tikzstyle{my-box}=[
rectangle,
draw=hidden-black,
rounded corners,
text opacity=1,
minimum height=1.5em,
minimum width=5em,
inner sep=2pt,
align=center,
fill opacity=.5,
]
\tikzstyle{leaf}=[
my-box, 
minimum height=1.5em,
fill=yellow!32, 
text=black,
align=left,
font=\normalsize,
inner xsep=2pt,
inner ysep=4pt,
]
\tikzstyle{leaf2}=[
my-box, 
minimum height=1.5em,
fill=purple!27, 
text=black,
align=left,
font=\normalsize,
inner xsep=2pt,
inner ysep=4pt,
]

\begin{figure*}[t]
\vspace{-2mm}
\centering
\resizebox{\textwidth}{!}{
	\begin{forest}
		forked edges,
		for tree={
			grow=east,
			reversed=true,
			anchor=base west,
			parent anchor=east,
			child anchor=west,
			base=left,
			font=\large,
			rectangle,
			draw=hidden-black,
			rounded corners,
			align=left,
			minimum width=4em,
			edge+={darkgray, line width=1pt},
			s sep=3pt,
			inner xsep=2pt,
			inner ysep=3pt,
			line width=0.8pt,
			ver/.style={rotate=90, child anchor=north, parent anchor=south, anchor=center},
		},
		where level=1{text width=14.7em,font=\normalsize,}{},
		where level=2{text width=8.0em,font=\normalsize,}{},
		where level=3{text width=9.5em,font=\normalsize,}{},
		where level=4{text width=12em,font=\normalsize,}{},
		[Mitigation Methods~(\S\ref{sec:mitigation}),ver
		[Parameter-Tuning Alignment~(\S\ref{subsec:parameter_tuning}),ver
			[Improving Training \\Scheme
                    [Contrastive \\Learning, text width=6em
                        [\eg~CIT~\cite{hu2023ciem}{,} HACL~\cite{jiang2024hallucination}
                        , leaf, text width=42em]
                    ]
                    [Instruction \\Tuning, text width=6em
                        [\eg~IDK-Instructions~\cite{cha2024visually}{,} REVERIE~\cite{zhang2025reflective}{,} TextSquare\\~\cite{tang2024textsquare}{,} LRV-Instruction~\cite{liu2023mitigating}{,} AIT~\cite{park2024mitigating}{,} MMINSTRUCT\\~\cite{liu2024mminstruct}
                        , leaf, text width=42em]
                    ]
                    [RLHF, text width=6em
                        [\eg~Fact-RLHF~\cite{sun2023aligning}{,} RLHF-V~\cite{yu2024rlhf}
                        , leaf, text width=42em]
                    ]
                    [Preference \\Optimization, text width=6em
                        [\eg~CLIP-DPO~\cite{ouali2025clip}{,} HA-DPO~\cite{zhao2023beyond}{,} FDPO~\cite{gunjal2024detecting}{,} \\HSA-DPO~\cite{xiao2024detecting}{,} HalluciDoctor~\cite{yu2024hallucidoctor}{,} \citet{chen2023mitigating}{,} ReCaption\\~\cite{wang2024mitigating2}{,} MOCHa~\cite{ben2024mitigating}{,} SILKIE~\cite{li2023silkie}{,} V-DPO\\~\cite{xie2024v}{,} RoVRM~\cite{wang2024rovrm}
                        , leaf, text width=42em]
                    ]
                ]
			[Improving Model \\Architecture
                    [Vision \\Encoder, text width=6em
                        [\eg~InternVL~\cite{chen2024internvl}{,} Ferret~\cite{you2023ferret}{,} Vcoder~\cite{jain2024vcoder}{,} Monkey\\~\cite{li2024monkey}{,} ObjMLM~\cite{dai2022plausible}
                        , leaf, text width=42em]
                    ]
                    [Connection \\Module, text width=6em
                        [\eg~PATCH~\cite{shang2024pixels}{,} HallE-Switch~\cite{zhai2023halle}
                        , leaf, text width=42em]
                    ]
                ]
		]
		[Parameter-Frozen Alignment~(\S\ref{subsec:parameter_frozen}),ver
			[Augment-Based
                    [Retrieval, text width=6em
                        [\eg~ARA~\cite{qu2024alleviating}{,} RAM~\cite{chen2023retrieval}{,} LMCap~\cite{ramos2023lmcap}{,} Smallcap\\~\cite{ramos2023smallcap}{,} \citet{sarto2024towards}{,} Pensieve~\cite{yang2024pensieve}
                        , leaf2, text width=42em]
                    ]
                    [Generate, text width=6em
                        [\eg~MARINE~\cite{zhao2024mitigating}{,} \citet{li2024visual}{,}  \citet{zhao2023enhancing}{,} RITUAL\\~\cite{woo2024ritual}{,} \citet{kim2024if}{,} VDGD~\cite{ghosh2024visual}
                        , leaf2, text width=42em]
                    ]
                ]
			[Inference-Based
                    [\eg~VTI~\cite{liu2024reducing}{,} ICT~\cite{chen2024ict}{,} MetaToken~\cite{fieback2024metatoken}
                    , leaf2, text width=49.6em]
                ]
			[Decoding-Based
                    [\eg~AVC~\cite{woo2024don}{,} DAMRO~\cite{gong2024damro}{,} VCD~\cite{leng2024mitigating}{,} ICD~\cite{wang2024mitigating}{,} \\CODE~\cite{kim2024code}{,} PAI~\cite{liu2407paying}{,} OPERA~\cite{huang2024opera}{,} SGD~\cite{min2024mitigating}{,} \citet{han2024skip}{,} \\\citet{yue2024less}{,} IBD~\cite{zhu2024ibd}{,} HALC~\cite{chen2024halc}{,} DBD~\cite{feng2024more}{,} AGLA\\~\cite{an2024agla}{,} RVD~\cite{zhong2024investigating}{,} M3ID~\cite{favero2024multi}{,} 
                    DeCo~\cite{wang2024mllm}{,} CGD\\~\cite{deng2024seeing}
                    , leaf2, text width=49.6em]
                ]
			[Post-Decoding
                    [\eg~LURE~\cite{zhou2023analyzing}{,} Woodpecker~\cite{yin2023woodpecker}{,} \citet{Yu_Jalaian_Bastian_2024}{,} LogicCheckGPT~\cite{wu2024logical}{,} \\VOLCANO~\cite{lee2023volcano}
                    , leaf2, text width=49.6em]
                ]
		]
		]
	\end{forest}
}
\vspace{-6mm}
\caption{Taxonomy of Misalignment Mitigation Methods for LVLMs, including \textit{Parameter-Tuning Alignment} and \textit{Parameter-Frozen Alignment}.}
\label{fig:taxonomy}
\vspace{-3mm}
\end{figure*}

\subsubsection{Inference Level}

Misalignment can also occur during the inference stage due to \emph{task discrepancy}. 
This discrepancy fundamentally represents an out-of-distribution (OOD) generalization problem, as users often pose questions or request tasks that deviate from the distribution of examples seen during training. Even when a LVLM has been trained on a large and diverse dataset, it may encounter novel combinations of visual and textual elements or be asked to perform tasks in ways that differ subtly but significantly from its training examples.
This OOD challenge manifests in several ways. First, the training data used for pre-training or fine-tuning the model may not fully align with the specific tasks it is later expected to perform~\cite{zhang2024rethinking}. For example, a model trained primarily on image captioning data might struggle when asked to answer specific questions about spatial relationships or perform detailed visual reasoning tasks. Second, users may phrase requests in ways that differ from the instruction patterns seen during training, leading to potential misinterpretation of the task requirements. Third, the visual inputs during inference may contain novel object configurations or scene compositions not well-represented in the training data. These distribution shifts can create misalignment in LVLMs as the model struggles to adapt to new and distinct tasks that require different interpretations of visual and textual information.

\section{Mitigation Methods}
\label{sec:mitigation}

Building upon our analysis of misalignment causes in LVLMs, we now examine strategies for mitigating these challenges (see Figure~\ref{fig:taxonomy}). These mitigation approaches can be categorized into two groups: parameter-tuning alignment methods and parameter-frozen alignment methods. Parameter-tuning alignment involves modifying specific components within the LVLM architecture to reduce misalignment through targeted parameter updates. In contrast, parameter-frozen alignment methods address misalignment while maintaining the LVLM's original parameters unchanged, offering solutions that preserve the model's structure while improving its cross-modal alignment capabilities. 

\subsection{Parameter Tuning Alignment}
\label{subsec:parameter_tuning}

Parameter-tuning alignment focuses on mitigating misalignment by refining the training scheme or enhancing the architecture itself.

\vspace{3pt}
\noindent\textbf{Improving Training Scheme.}\,
Parameter-tuning methods that improve the training scheme often address misalignment broadly as a data-level issue or as a general visual-textual misalignment \cite{ouali2025clip, jiang2024hallucination}. This understanding leads to a straightforward objective, which is reducing the modality gap between visual and textual representations. This can often achieved by improving the dataset quality or optimizing training techniques. One common approach is contrastive learning, exemplified by methods such as CIT \cite{hu2023ciem} and HACL \cite{jiang2024hallucination}. These techniques involve using a third model to generate positive and negative data pairs. The LVLM is then trained to bring the representations of positive pairs closer together while pushing negative pairs apart in the embedding space. Another widely adopted strategy is instruction tuning, as seen in LRV-Instruction \cite{liu2023mitigating}, VideoCon \cite{bansal2024videocon}, HACA \cite{zhao2025can}, and TextSquare \cite{tang2024textsquare}. Similarly, these approaches rely on a third model to generate instructional data, which is subsequently used to train the LVLM effectively. However, these approaches often lack robust quality assurance mechanisms to verify the accuracy or relevance of the generated data, introducing potential risks. Alternatively, Reinforcement Learning from Human Feedback (RLHF) employs human feedback to train a reward model, ensuring that the generated data aligns with human preferences \cite{sun2023aligning, yu2024rlhf}. While RLHF guarantees high-quality training data, it comes at a significant cost. To address this, some methods leverage preference optimization, wherein multiple responses are generated for the same input image, ranked or scored by a third model, and categorized into positive and negative pairs \cite{ouali2025clip, zhao2023beyond, gunjal2024detecting}. The model is then fine-tuned on this curated dataset. Although these methods can significantly improve the model, they are often constrained by either high resource requirements (as in RLHF) or the uncertain quality of generated data (as in contrastive learning and instruction tuning) or rerank model (as in preference optimization). This highlights the ongoing need for large, diverse, and high-quality datasets to effectively address data-level misalignment.

\vspace{3pt}
\noindent\textbf{Improving Model Architecture.}\,
Methods that improve the model architecture often involve a deep understanding of the root causes of misalignment, allowing researchers to pinpoint deficiencies within specific components of the LVLM. Typical LVLM architectures consist of three main components: the visual encoder, the adapter module, and the LLM \cite{liu2024survey, bai2024hallucination}. Most architecture-focused approaches concentrate on enhancing the visual encoder or the adapter module, with relatively few addressing improvements to the LLM itself. This aligns with our earlier model-level claim of the model ability gap, where the LLM often outperforms the visual encoder. Blindly enhancing the LLM could exacerbate this gap, potentially worsening the misalignment issue. To reduce this ability gap, some studies scale up the visual encoder by increasing its parameter size \cite{chen2024internvl}. Others introduce additional components to the visual encoder to improve its capabilities without necessarily scaling up its size \cite{you2023ferret, jain2024vcoder, li2024monkey}. In addition to the visual encoder, many methods focus on improving the adapter module, which serves as the critical bridge between the visual and textual modalities. Enhancements to the adapter module often involve adding intermediary layers or mechanisms to better align the visual encoder's outputs with the LLM's input requirements. For example, PATCH \cite{shang2024pixels} employs trainable virtual tokens to enhance the projection layer, improving cross-modal alignment. Similarly, HallE-Switch \cite{zhai2023halle} introduces a dynamic mechanism that adjusts the flow of information between the visual encoder and the LLM based on input complexity. By addressing these architectural components, parameter-tuning methods aim to reduce the modality gap and improve the alignment between visual and textual representations, ultimately enhancing the LVLM's performance across tasks.

\renewcommand{\arraystretch}{0.8}
\setlength{\tabcolsep}{3pt}
\begin{table*}[h]
    \caption{Comparison of Mitigation Methods on the POPE Benchmark. The table demonstrates the performance of the baseline model and different mitigation methods that address baseline misalignment. Performance is evaluated using Accuracy, Precision, Recall, and F1 score (with the highest score in each category underlined). Additionally, the computational cost (in seconds per input) for both the baseline and each mitigation method is provided.}
    \centering
    \small
    \scalebox{0.92}{
    \begin{tabular}{l 
                    *{4}{c} | 
                    *{4}{c} | 
                    *{4}{c} | 
                    c}
        \toprule
        & \multicolumn{4}{c|}{\textbf{POPE-Random}} 
        & \multicolumn{4}{c|}{\textbf{POPE-Adversarial}} 
        & \multicolumn{4}{c|}{\textbf{POPE-Popular}} 
        & \textbf{Comp. Cost} \\
        \cmidrule(lr){2-5}\cmidrule(lr){6-9}\cmidrule(lr){10-13}\cmidrule(lr){14-14}
        \textbf{Method} & Acc & Precis & Recall & F1 
        & Acc & Precis & Recall & F1 
        & Acc & Precis & Recall & F1 
        & Sec/Input \\
        \midrule
        \textbf{Baseline} \\
        LLaVa-V1.5-7B 
        & 88.4 & 87.5 & 89.7 & 88.6 
        & 77.9 & 72.6 & 89.7 & 80.3 
        & 84.8 & 81.7 & 89.7 & 85.5 
        & 1.27 \\[1ex]
        \midrule
        \textbf{Mitigations} \\
        SoM-LLaVA \cite{yan2024list} 
        & \underline{89.6} & 89.1 & \underline{90.2} & \underline{89.6} 
        & 81.0 & 76.2 & \underline{90.2} & 82.6 
        & 84.8 & 81.4 & \underline{90.2} & 85.6 
        & 1.62 \\[1ex]
        SID \cite{huo2024self} 
        & 89.4 & 92.2 & 86.1 & 89.0 
        & 80.4 & 77.4 & 85.9 & 81.4 
        & 85.9 & 85.9 & 85.9 & \underline{85.9} 
        & 1.68 \\[1ex]
        LogicCheckGPT \cite{wu2024logical} 
        & 88.0 & \underline{98.3} & 77.3 & 86.5 
        & \underline{85.0} & \underline{94.1} & 74.7 & \underline{83.3} 
        & \underline{86.7} & \underline{95.1} & 77.3 & 85.3 
        & 51.5 \\
        \bottomrule
    \end{tabular}}
    \label{tab:mitigation_comparison}
\end{table*}

\subsection{Parameter Frozen Alignment}
\label{subsec:parameter_frozen}

Parameter-frozen alignment methods have gained increasing popularity due to their significant practical advantages. These training-free approaches are highly modular and easy to implement, allowing them to be readily integrated into existing systems without requiring costly retraining or fine-tuning processes. This makes them particularly attractive for real-world applications where computational resources may be limited. We categorize these parameter-frozen methods into four types based on where they intervene in the LVLM processing pipeline: Augment-based mitigation, augmenting the LVLM by incorporating external knowledge; inference-based mitigation, operating in the model's latent space during intermediate processing; decoding-based mitigation, which guides the text generation process; and post-decoding mitigation, which refines the final outputs.

\vspace{3pt}
\noindent\textbf{Augment-based Methods.}\,
As analyzed in Section 3, insufficient input of image information is one of the primary causes of misalignment, leading to poor visual understanding. To address this, retrieval-augmented generation (RAG) methods have been adapted to dynamically integrate external knowledge into LVLMs through retrieved results \cite{qu2024alleviating, chen2023retrieval, ramos2023lmcap, ramos2023smallcap, sarto2024towards, yang2024pensieve}. By reranking the similarity of image-text pairs, RAG approaches provide more visual context and guidance to the model. Similarly, other methods rely on generating approach to enrich the input with additional information. For instance, \citet{zhao2024mitigating}, \citet{li2024visual}, and \citet{zhao2023enhancing} propose integrating an auxiliary model to generate relevant information based on the image. Then inject these information into the input prompt. Alternatively, methods such as RITUAL \cite{woo2024ritual} bypass the need for external models. It enhances the model's exposure to diverse visual contexts by applying random transformations to input images. Additionally, approaches like \cite{kim2024if, ghosh2024visual} employ self-generated textual descriptions appended to the input prompt, ensuring the model has sufficient knowledge to answer questions accurately.

\vspace{3pt}
\noindent\textbf{Inference-based Methods.}\,
Some methods operate in the model's latent space during the inference process, prior to decoding, by intervening in both visual and textual representations to improve alignment. For instance, Visual and Textual Intervention (VTI) \cite{liu2024reducing} pre-computes intervention directions using a small set of examples and applies them during inference to enhance feature stability and vision-text alignment, without requiring additional training. Similarly, Image-Object Cross-Level Trusted Intervention (ICT) \cite{chen2024ict} introduces a lightweight mechanism that intervenes in the model's attention at both image and object levels, applying targeted activation shifts to selected attention heads. Since they operate directly on the model's internal representations, they can make precise adjustments to improve alignment without disrupting the model's broader language understanding capabilities. This makes inference-based methods effective at reducing misalignment while preserving the model's ability to generate contextually appropriate responses.

\vspace{3pt}
\noindent\textbf{Decoding-based Methods.}\,
Another common approach to mitigating misalignment involves modifying the decoding process. These methods often target issues of imbalanced attention. However, the imbalance attention between what still remain debated. Some researchers argue that the model over-focuses on irrelevant image tokens, such as background elements or unimportant details \cite{woo2024don, gong2024damro}. However, the prevailing view is that the model prioritizes textual tokens over visual ones, neglecting critical visual information \cite{leng2024mitigating, wang2024mitigating, kim2024code, liu2407paying}. Despite these differences in interpretation, most decoding-based methods use contrastive decoding to rebalance attention between modalities, typically by reducing attention to textual tokens while enhancing focus on visual tokens. This approach, however, contrasts with inference-based methods, which avoid reducing attention to textual information and instead preserve the model's overall language understanding. Another interesting observation is that, while decoding-based methods typically lead to similar approaches, they can sometimes result in divergent strategies. For instance, OPERA \cite{huang2024opera} hypothesizes that the model over-relies on summary tokens, instead of focusing visual tokens. However, text summarization is SGD's solution \cite{min2024mitigating} to misalignment. It uses summarization to shorten textual context and helps model shift focus toward visual information. This divergence underscores how subtle differences in understanding misalignment's root causes can lead to contradicted methodologies.

\vspace{3pt}
\noindent\textbf{Post-decoding Methods.}\,
Lastly, post-decoding approaches present broader hypotheses about misalignment causes, tackling issues ranging from data-level biases to model-level deficiencies. Methods such as LURE \cite{zhou2023analyzing} and Woodpecker \cite{yin2023woodpecker} exemplify this category. LURE focuses on addressing object hallucinations by revising the generated text, identifying hallucinatory content, and reconstructing less biased outputs. Woodpecker employs a five stages validation mechanism to extract and correct inconsistencies in the generated response. Despite their specific details, these methods converge on a shared strategy, which involves modifying the model's outputs after decoding without altering parameters or architecture, making them easily adaptable to various LVLMs. This flexibility lies in their goal-oriented nature, as they directly target specific misalignment phenomena. 

\subsection{Mitigation Performance Comparison}
As shown in Table \ref{tab:mitigation_comparison}, we selected LLaVa-V1.5-7B as the baseline model and evaluated it on POPE benchmarks. We then tested three mitigation methods: SoM-LLaVA \cite{yan2024list}, SID \cite{huo2024self}, and LogicCheckGPT \cite{wu2024logical}, to assess their effectiveness and efficiency in mitigating misalignment in the baseline model. The baseline model was loaded directly from Hugging Face, with parameters in float16 precision and all other implementation details set to default. Each mitigation method was implemented strictly according to its official GitHub repository, without modifications to core components, except for necessary adjustments such as path configurations. More detailed analysis regarding the mitigation performance please refer to Section~\ref{appendix:mitigation_comparison} in the Appendix.

\section{Conclusions}

In this paper, we systematically survey alignment and misalignment in LVLMs through an explainability lens. We show that achieving proper alignment involves complex interactions between data quality, model architecture, and inference procedures. We categorize misalignment into object, attribute, and relational levels, providing a clear framework to understand these challenges and develop targeted solutions.
Our review of mitigation strategies highlights approaches ranging from computationally intensive parameter-tuning methods to practical parameter-frozen solutions, each with trade-offs in effectiveness and feasibility. Lastly, we outline key future directions for building robust vision-language systems (see Appendix \ref{appendix:future}).

\section*{Limitations}
While this paper provides a comprehensive survey of alignment and misalignment in LVLMs, we acknowledged there are several limitations. Currently, the scope of this survey is limited to the alignment between vision and language modalities. However, real-world applications often require the integration of multiple modalities, such as audio, video, or sensor data. We plan to expand the scope to explore the challenges and alignment techniques in such complex multi-modal settings, including investigating alignment mechanisms in architectures that incorporate three or more modalities, developing new evaluation metrics for multi-modal alignment, and studying the interactions between different modalities in integrated systems.

\section*{Acknowledgments}
 Mengnan Du is in part supported by National Science Foundation (NSF) Grant \#2310261. 
 Lu Cheng is supported by the National Science Foundation (NSF) Grant \#2312862, NSF CAREER \#2440542, NSF-Simons SkAI Institute, National Institutes of Health (NIH) \#R01AG091762, Google Research Scholar Award, and a Cisco gift grant.

\bibliography{acl_latex}

\clearpage

\appendix

\begin{figure*}
    \centering
    \includegraphics[width=1\linewidth]{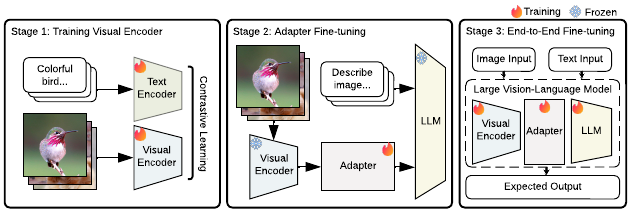}
    \caption{Overview of the three-stage LVLM training process, showing the progression from contrastive learning of visual-text encoders, through adapter fine-tuning with frozen components, to end-to-end model training.}
    \label{fig:aligment_procedure}
\end{figure*}

\begin{table*}[t]
\centering
\caption{Comparison of Vision-Language Models}
\scalebox{0.75}{  
\begin{tabular}{lllll}
\toprule
& \textbf{Vision Encoder} & \textbf{Adapter} & \textbf{LLM} \\
\midrule



\textbf{Qwen-VL} \cite{Qwen-VL} & Vision Transformer (ViT) & Single cross-attention layer & Qwen \\

\textbf{Qwen2-VL} \cite{wang2024qwen2} & ViT & MLP layer & Qwen2 \\

\textbf{MiniGPT-4} \cite{zhu2023minigpt} & ViT-G/14 & Single linear projection layer & Vicuna \\

\textbf{Llama 3.2-Vision} \cite{meta2024llama32} & Modified ViT (16×16 patches) & Multiple cross-attention layers & Llama 3.1 text-only model \\

\textbf{LLAVA-1.5} \cite{liu2024visual} & CLIP-ViT-L-336px & MLP projection layer & Vicuna v1.5 \\

\textbf{DeepSeek-VL2} \cite{wu2024deepseekvl2mixtureofexpertsvisionlanguagemodels} & SigLIP-SO400M-384 & Two-layer MLP & DeepSeekMoE \\

\bottomrule
\end{tabular}
}
\label{tab:vlmodels}
\end{table*}

\section{Overview of LVLM Architectures}
\label{appendix:architectures}

Table 1 summarizes the architectural components of six prominent LVLMs discussed throughout this survey paper on model alignment and misalignment. The comparison reveals common patterns in LVLM design: transformer-based vision encoders (predominantly ViT variants), adapter modules of varying complexity to connect vision and language components, and state-of-the-art language models.

\subsection{How is Alignment Achieved?}

The development of alignment in LVLMs progresses through three major stages (see Figure~\ref{fig:aligment_procedure}), each is built upon its predecessor to achieve increasingly sophisticated cross-modal integration. Additionally, we present prominent real-world LVLM architectures in Appendix \ref{appendix:architectures}.

\vspace{3pt}
\noindent\textbf{Stage 1: Training Visual Encoders.}\,
The foundation of LVLM alignment begins with training visual encoders through contrastive learning, exemplified by models like CLIP \cite{radford2021learning}. In this stage, the model learns to align visual and textual representations in a shared embedding space through a contrastive loss function. The process involves training on large-scale image-text pairs where matching pairs are pulled together in the embedding space while non-matching pairs are pushed apart. This leads to the development of robust visual representations that can meaningfully correspond to textual descriptions. Through this process, a visual encoder is created that can extract semantically meaningful features from images in a way that naturally aligns with language. This initial stage is crucial as it establishes the basic capability for cross-modal understanding, though the alignment is still relatively coarse-grained.

\vspace{3pt}
\noindent\textbf{Stage 2: Adapter Fine-tuning.}\,
The second stage involves fine-tuning an adapter module that bridges the pre-trained visual encoder with the language model \cite{yang2024mma}. This stage introduces lightweight adapter architectures, which typically consist of simple components such as linear layers, MLPs, or cross-attention layers that learn to translate between visual and language model embedding spaces. For example, cross-attention layers can feed image encoder representations into the language model, enabling the model to attend to relevant visual features when generating text~\cite{meta2024llama32}. A key characteristic of this approach is the preservation of the original capabilities of both the visual encoder and language model while learning to interface between them. During adapter training, while the visual encoder parameters may be updated, the language model parameters often remain frozen to maintain their original text capabilities.  This intermediate stage is essential for establishing effective connections between modalities while preserving the specialized capabilities of each component.

\vspace{3pt}
\noindent\textbf{Stage 3: End-to-End Fine-tuning.}\,
The final stage involves comprehensive fine-tuning of the entire system, including the visual encoder, adapter, and LLM components together \cite{zhai2024fine}. This comprehensive approach allows for deeper integration and more sophisticated alignment between all components. It enables the model to learn task-specific optimizations that require coordinated adjustments across all modules. Through this process, the model develops more advanced cross-modal understanding capabilities and facilitates the emergence of emergent behaviors that arise from the deep integration of visual and textual processing. This stage results in the highest performance but requires careful balancing to avoid catastrophic forgetting or degradation of pre-existing capabilities.

\section{Representation Misalignment}
\label{appendix:representation_misalignment}
At the representation level, alignment refers to how closely visual representations $v \in \mathcal{V}$ and textual representations $t \in \mathcal{T}$ are mapped within the shared embedding space $\mathcal{E}$ of an LVLM. Representation misalignment occurs when semantically corresponding visual and textual representations occupy distant regions in this space, resulting in low similarity. This misalignment can have significant downstream consequences. For example, if the embedding of an image containing an apple lies far from the embedding of the word "apple" and is instead closer to the embedding of "dog", the model may incorrectly interpret the image as depicting a dog. Such misalignment at the representation level undermines the model's ability to ground visual input in the correct textual context. In this paper, we argue that representation misalignment fundamentally determines the upper ceiling of an LVLM’s performance, regardless of the quality of its language modeling or vision encoder in isolation.

\section{How to Measure Alignment?}
\label{appendix:measure_alignment}

This section examines approaches for quantifying the effectiveness of alignment in LVLMs. These measurement approaches naturally align with our earlier definition in Section 2.1 of representation alignment and behavioral alignment, and can be organized along these two fundamental levels.

\vspace{3pt}
\noindent\textbf{Representation Level.}\,
At the representation level, alignment can be directly measured between visual and textual representations within the LVLM's embedding space by assessing how similarly the visual and textual modalities encode and relate to the same concepts or data points. The simplest approach is to compute the cosine similarity between the embeddings of visual and textual data. High alignment corresponds to scores close to 1, while low alignment corresponds to scores closer to 0 \cite{shu2024exploring}. More sophisticated metrics have been developed to assess alignment between the two representation spaces. For instance, the mutual nearest-neighbor metric quantifies alignment by evaluating the consistency of nearest neighbors across modalities \cite{huh2024platonic}. Another approach is kernel alignment, which evaluates the similarity of pairwise relationships within each modality's embedding space, providing a holistic view of the alignment structure \cite{maniparambil2024vision}.

\vspace{3pt}
\noindent\textbf{Behavioral Level.}\,
The behavioral level measures alignment through the model's performance on various downstream tasks and benchmarks, using both direct comparisons and automated evaluation systems. The strength of alignment directly impacts the LVLM's performance, as better alignment typically leads to improved task outcomes. These measurements generally involve comparing the model's outputs against ground truth labels, either through direct comparison or using evaluation models to simulate human judgment. Numerous benchmarks have been developed to assess LVLM alignment across a range of tasks, from coarse-grained evaluations (e.g., object existence) to fine-grained assessments (e.g., color, count, spatial relations). Examples of such benchmarks include POPE \cite{li2023evaluating}, CHAIR \cite{rohrbach2018object}, MME \cite{fu2023mme}, MMHal-Bench \cite{sun2023aligning}, and LLaVa-Bench \cite{liu2024visual}. In addition to traditional benchmarks, advanced evaluation models like GAVIE \cite{liu2023mitigating}, CCEval \cite{zhai2023halle} and HaELM \cite{wang2023evaluation} provide sophisticated assessments by considering context and evaluating responses comprehensively, similar to human evaluators. The flexibility and diversity of evaluation models enable thorough measurement capabilities needed for open-ended questions. 

In Table \ref{tab:benchmark}, we compare several existing benchmarks. While most of these benchmarks focus on evaluating misalignment, some also assess reasoning capabilities, such as whether the LVLM can reason based on image information or whether the LVLM fully follow user input instructions. Future benchmarks should also adopt this comprehensive approach, evaluating not only misalignment but also whether mitigating misalignment impacts the LVLM's reasoning ability, as our goal is to develop LVLMs with minimal misalignment while preserving their reasoning capabilities.

\section{Formal Theoretical Perspective}
\label{appendix:theoretical_perspective}

From a theoretical standpoint, visual and textual modalities can be modeled as distinct projections of a shared latent semantic space \cite{ngiam2011multimodal}. Let $Z \in \mathcal{Z}$ denote the latent variable representing the underlying world state or concept being observed. Images $x_v \in \mathcal{X}_v$ and text $x_t \in \mathcal{X}_t$ are generated via distinct, conditionally independent observation channels:
\begin{equation}
    x_v \sim p_v(x \mid Z), \quad x_t \sim p_t(x \mid Z).
\end{equation}
We assume visual and textual data are conditionally independent given the latent semantics:
\begin{equation}
    p(x_v, x_t \mid Z) = p_v(x_v \mid Z) \cdot p_t(x_t \mid Z).
\end{equation}
This assumption reflects the intuitive idea that both images and texts encode complementary but overlapping information about the same underlying concept, such as an object, attribute, or relational.

Let $f_v: \mathcal{X}_v \rightarrow \mathbb{R}^d$ and $f_t: \mathcal{X}_t \rightarrow \mathbb{R}^d$ denote the learned encoders (e.g., a vision encoder and a LLM) that map raw inputs to $d$-dimensional embeddings. If both $f_v(x_v)$ and $f_t(x_t)$ preserve all the relevant information about the latent variable $Z$, they are said to be sufficient statistics for $Z$:
\begin{align}
    I(Z; f_v(x_v)) = I(Z; x_v)\\
    I(Z; f_t(x_t)) = I(Z; x_t),
\end{align}
where symbol $I(A;B)$ denotes mutual information between variables $A$ and $B$. This implies that no semantic information about $Z$ is lost in the representation process \cite{tishby2015deep, poole2019variational}. Consequently, the learned representations $f_v(x_v)$ and $f_t(x_t)$ should be functionally equivalent in terms of semantic content, as they reflect the same underlying meaning. This sets the foundation for representational alignment.

\section{More Reasons for Misalignment}
\label{appendix:more_reason}

Beyond the dataset-level issues discussed in Section~\ref{sec:data-level-issue}, we present additional dataset-related challenges that may contribute to the misalignment of LVLMs.

\begin{itemize}[leftmargin=*]\setlength\itemsep{-0.3em}

\item
\emph{Data Imbalance}: When certain classes or types of data are disproportionately represented, it skews the model’s training process \cite{liu2023mitigating, hu2023ciem}. For example, visual question-answering datasets often overrepresent positive answers, subtly training the model to favor these outcomes while underperforming on underrepresented negative answers. 

\item 
\emph{Data Inconsistency}:
Inconsistencies exacerbate misalignment by introducing contradictory outputs across different tasks for the same image. For instance, an image captioning task might describe an image as depicting `a tiger eating a chicken,’ yet in a visual question-answering task for the same image, the answer to `what is the tiger eating?’ might label the prey as `a duck’ \cite{maharana2023exposing}. Such contradictions disrupt the model’s ability to generate coherent and consistent outputs across tasks.

\item 
\emph{Data False Negative}:
False negatives in the dataset further complicate alignment, as negative image-text pairs, though not perfectly matching, share overlapping components \cite{liang2022mind, byun2024mafa}. During training, embeddings of positive pairs are drawn closer together, while those of negative pairs are pushed apart. This binary method can suppress latent similarities within false negatives, reducing the model’s capacity to effectively align diverse modalities. 

\item
\emph{Data Polysemy}:
The inherent polysemy within datasets introduces additional complexity. Polysemy enriches data diversity by allowing a single word or image to convey multiple meanings depending on context, but this ambiguity also amplifies the risk of misalignment \cite{ma2020addressing, dingimp}. For example, an image caption of `the bat hit the ball' could refer to the animal or the baseball bat. This variability challenges the model to establish consistent mappings between modalities. 
\end{itemize}



\renewcommand{\arraystretch}{0.6}  
\renewcommand\cellgape{}  

\begin{table*}[t]
\centering
\caption{Comparison of Misalignment Benchmark across Different Evaluation Dimensions. The evaluation dimensions are categorized into two main aspects: (1) Evaluating misalignment, which examines object, attribute, and relational misalignments in model outputs; and (2) Evaluating reasoning, which assesses models' cognitive reasoning abilities and instruction following capabilities. The rightmost column indicates whether the benchmark uses traditional evaluation methods or third-model evaluation approaches.}
\scalebox{0.77}{  
\begin{tabular}{l c c c c c c}
\toprule
& \multicolumn{3}{c}{\textbf{Evaluating Misalignment}} & \multicolumn{2}{c}{\textbf{Evaluating Reasoning}} & \\
\cmidrule(lr){2-4} \cmidrule(lr){5-6} \addlinespace[-2pt] 

& \shortstack{\textbf{Object} \\ \textbf{Misalignment}} 
& \shortstack{\textbf{Attribute} \\ \textbf{Misalignment}} 
& \shortstack{\textbf{Relational} \\ \textbf{Misalignment}}  
& \shortstack{\textbf{Cognition} \\ \textbf{Reasoning}} 
& \shortstack{\textbf{Instruction} \\ \textbf{Following}} 
& \shortstack{\textbf{Traditional/} \\ \textbf{Third-model}} \\
\midrule
\textbf{POPE} \cite{li2023evaluating} & \cmark & \xmark & \xmark & \xmark & \xmark & Traditional \\

\textbf{CHAIR} \cite{rohrbach2018object} & \cmark & \xmark & \xmark & \xmark & \xmark & Traditional \\

\textbf{MME} \cite{fu2023mme}  & \cmark & \cmark & \cmark & \cmark & \xmark & Traditional \\

\textbf{MMHal-Bench} \cite{sun2023aligning} & \cmark & \cmark & \cmark & \cmark & \xmark & Traditional \\

\textbf{LLaVa-Bench} \cite{liu2024visual} & \cmark & \cmark & \cmark & \xmark & \xmark & Traditional \\

\textbf{LVLM-eHub} \cite{xu2024lvlm} & \cmark & \cmark & \xmark & \cmark & \xmark & Traditional \\

\textbf{GAVIE} \cite{liu2023mitigating} & \cmark & \cmark & \cmark & \xmark & \cmark & Third-model \\

\textbf{CCEval} \cite{zhai2023halle} & \cmark & \xmark & \xmark & \xmark & \xmark & Third-model \\

\textbf{HaELM} \cite{wang2023evaluation} & \cmark & \cmark & \cmark & \xmark & \xmark & Third-model \\

\bottomrule
\end{tabular}
}
\label{tab:benchmark}
\end{table*}

\section{Examples of LVLM Descriptions}
\label{appendix:example}

\begin{figure*}
    \centering
    \includegraphics[width=1\linewidth]{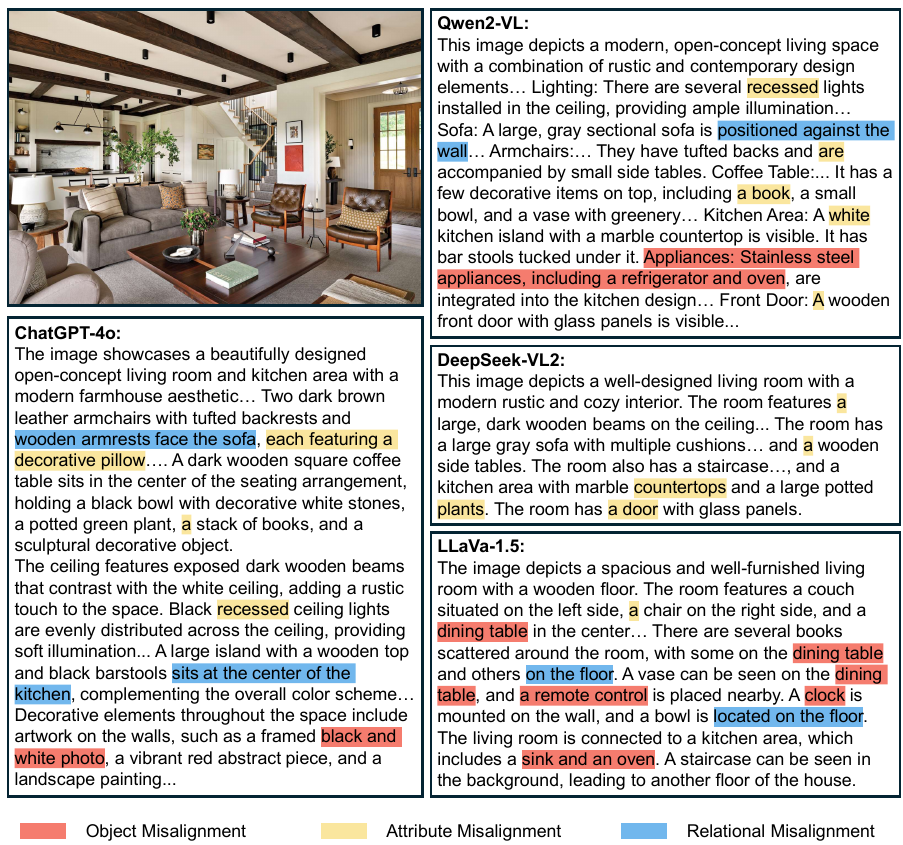}
    \caption{This figure presents examples of descriptions generated by different LVLMs when given the image shown in the upper left corner. The highlighted text segments in the descriptions represent different types of misalignment. Specifically, red highlights indicate object misalignment. Yellow highlights represent attribute misalignment. Blue highlights denote relational misalignment.}
    \label{fig:example_1}
\end{figure*}

\begin{figure*}
    \centering
    \includegraphics[width=1\linewidth]{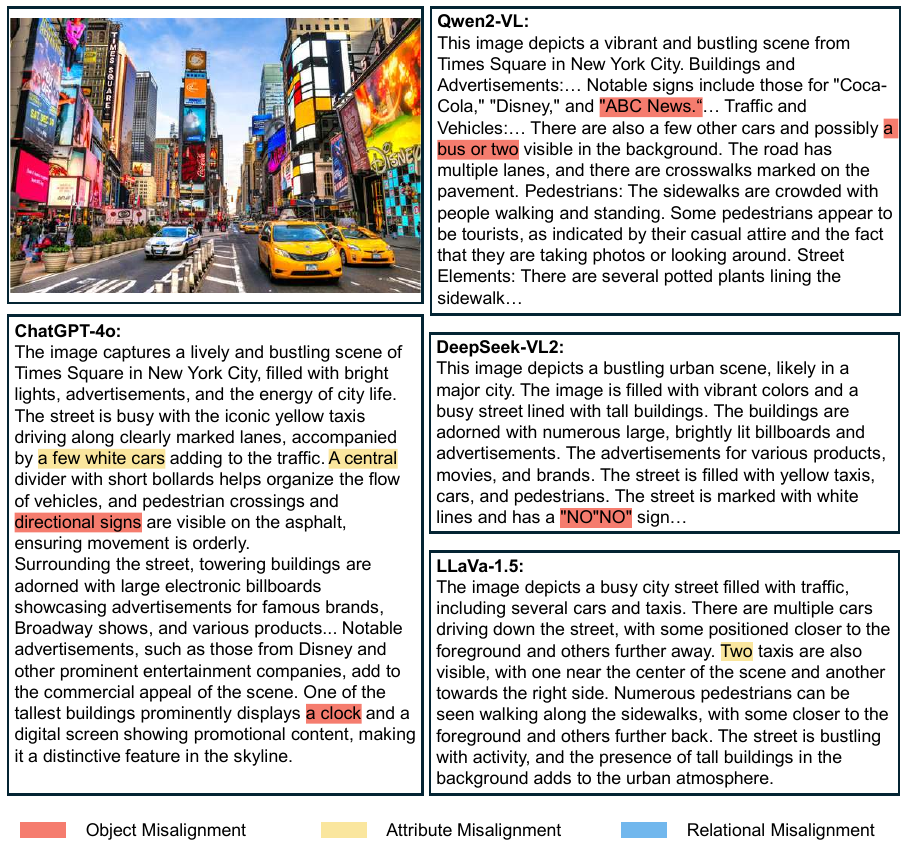}
    \caption{This figure presents examples of descriptions generated by different LVLMs when given the image shown in the upper left corner. The highlighted text segments in the descriptions represent different types of misalignment. Specifically, red highlights indicate object misalignment. Yellow highlights represent attribute misalignment. Blue highlights denote relational misalignment.}
    \label{fig:example_2}
\end{figure*}

\begin{figure*}
    \centering
    \includegraphics[width=1\linewidth]{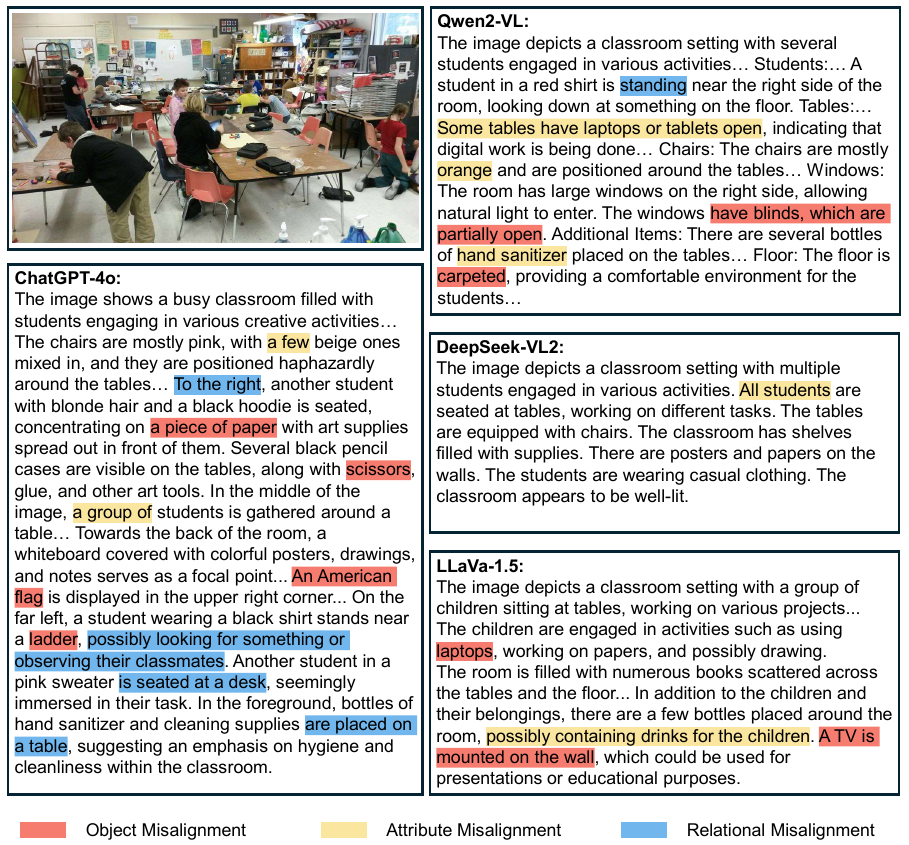}
    \caption{This figure presents examples of descriptions generated by different LVLMs when given the image shown in the upper left corner. The highlighted text segments in the descriptions represent different types of misalignment. Specifically, red highlights indicate object misalignment. Yellow highlights represent attribute misalignment. Blue highlights denote relational misalignment.}
    \label{fig:example_3}
\end{figure*}

In this section, we present examples from four leading LVLMs and illustrate their generated descriptions for given images. We have selected one closed-source model, ChatGPT-4o, accessed through its official website \url{https://chatgpt.com/}, and three open-source models, Qwen2-VL-72B-Instruct, DeepSeek-VL2, and LLaVa-1.5-13B-hf, loaded directly from Hugging Face. All implementation settings, including parameters such as temperature, were kept at their default values to ensure consistency. The experiments were conducted using an A100 PCIE 80GB GPU. To enhance computational efficiency and performance, the models were loaded in float16 precision. During the evaluation phase, each model was provided with an image along with the following prompt: `Describe this image in detail. Include specific objects, their attributes, and relationships.' The input images and the corresponding descriptions generated by each model are presented in Figure \ref{fig:example_1}, \ref{fig:example_2}, and \ref{fig:example_3}.

\section{Detailed Analysis of Mitigation Methods Comparison}
\label{appendix:mitigation_comparison}

As shown in Table \ref{tab:mitigation_comparison}, we evaluated model performance using Accuracy, Precision, Recall, and F1-score, with the highest values underlined in the table. The results indicate that mitigation methods generally improve performance across most metrics compared to the baseline, particularly SoM-LLaVA and LogicCheckGPT, which achieved five and six highest scores, respectively. However, a deeper analysis reveals that some methods negatively impacted certain metrics. For instance, while LogicCheckGPT achieved the highest Precision on all benchmarks, it exhibited relatively lower Recall and F1 scores, sometimes even worse than the baseline. This occurs because LogicCheckGPT is conservative in predicting positives, meaning it only classifies a sample as positive when it is highly confident, resulting in fewer false positives but more false negatives. Interestingly, SoM-LLaVA consistently achieved the highest Recall across all benchmarks, indicating that it is less strict in determining positive cases compared to LogicCheckGPT. This highlights how different mitigation strategies lead to different prediction behaviors, emphasizing the need for future works to report performance across multiple metrics rather than focusing solely on a single high-performing metric.

Beyond performance, we also evaluated the computational cost of each mitigation method in seconds per input. While SoM-LLaVA and SID had only a minor increase in inference time compared to the baseline, LogicCheckGPT required significantly more time. This is because LogicCheckGPT relies on an external third model, such as GPT-3.5, to assess logical consistency and mitigate object misalignment, making its computational cost highly dependent on the third model’s complexity. This trade-off between performance and computational efficiency is critical, as high computational time may hinder real-world deployment of mitigation techniques. Future research should aim to balance effectiveness with efficiency to ensure practical usability.


\section{Future Research Directions}
\label{appendix:future}

In this section, we discuss several important directions for future research in understanding and improving alignment in LVLMs.

\subsection{Standardized Benchmarks}
The current evaluation of misalignment in LVLMs suffers from a critical limitation, i.e., the lack of standardized, comprehensive benchmarks that can systematically assess different types of misalignment across models. While existing benchmarks have made important contributions, they typically focus on specific aspects of misalignment in isolation. For instance, POPE~\cite{li2023evaluating} primarily evaluates object hallucination, while other benchmarks concentrate on particular relationship errors or attribute inconsistencies. What is urgently needed is a unified evaluation framework that can systematically assess misalignment across all semantic levels, from object-level (e.g., describing a non-existent dog in an image) to attribute-level (e.g., color, size, texture errors) and relation-level misalignment (e.g., spatial relationship errors). Such a comprehensive benchmark would enable direct comparisons between different LVLM architectures and alignment techniques using standardized metrics, evaluate both representational alignment and behavioral alignment, and assess how misalignment manifests across different types of tasks. The benchmark should also consider both the frequency and severity of different types of misalignment, rather than treating all misalignments as equally problematic. The development of such standardized benchmarks would represent a significant step forward in our understanding of misalignment in LVLMs and accelerate progress toward more reliable and trustworthy vision-language systems.

\subsection{Explainability based Diagnose}
To better understand and address alignment issues in LVLMs, future research should leverage advanced explainability techniques that can reveal the internal mechanisms of these models. There are two critical categories of explainability approaches that warrant investigation: (1) internal knowledge decoding and (2) attribution methods.

The first category of explainability approaches centers on internal knowledge decoding and understanding how information is processed within LVLMs~\cite{zhao2024opening,zhao2024explainability}. Mechanistic interpretability approaches could help identify specific components and circuits within LVLMs that are responsible for cross-modal alignment, providing insights into how visual and language representations are integrated and processed. Similarly, probing techniques can analyze the emergence and evolution of aligned representations across different layers and attention heads, helping researchers understand where and how misalignment occurs within the model architecture~\cite{zhao2024beyond}. This detailed understanding of the internal working mechanisms would not only advance theoretical knowledge but also guide the development of more effective alignment techniques.

The second critical category focuses on attribution methods that can determine the relative influence of different information sources on model outputs. LVLMs have three primary information sources for generating outputs: user text prompts, input images, and knowledge stored within pre-trained LLMs. Future research needs to develop sophisticated attribution algorithms that can determine whether a model's output primarily depends on the input text prompt, derives from the visual information in the image, or relies on the LLM's internal knowledge. This detailed attribution analysis would help identify when and why misalignment occurs, such as cases where the model inappropriately relies on LLM knowledge rather than visual evidence, or when it fails to properly integrate information from multiple sources. Such insights would be useful for designing targeted mitigation strategies that address specific types of misalignment and improve the overall reliability of LVLMs.

\subsection{Architectural Innovations}
Current LVLM architectures face fundamental challenges, including significant ability gaps between visual encoders and LLMs, persistent attention imbalances between modalities, and knowledge conflicts between visual and textual representations. While most existing solutions focus on improving training procedures or adding post-processing steps, future research should focus on architectural innovations that address these structural limitations. This could include developing novel integration mechanisms that better balance the capabilities of visual and language components, and dynamic architectures that can adaptively adjust their attention mechanisms to maintain equilibrium between modalities. The field would benefit from multi-stage processing architectures that explicitly manage knowledge conflicts through specialized components for different levels of semantic understanding. Additionally, new transformer architectures specifically designed for vision-language tasks, rather than adapted from unimodal architectures, could help bridge the ability gap between visual and textual processing. 

\subsection{Real-world Implications}
While LVLM misalignment mitigation strategies have achieved impressive improvements, most of these improvements are primarily validated on predefined benchmarks, with limited translation to real-world applications such as autonomous driving, medical imaging, or vision-focused assistive AI. However, different types of misalignment can have serious consequences in these domains. For instance, in autonomous driving, object or relational misalignment may cause the system to overlook pedestrians or misinterpret traffic scenarios, leading to safety-critical failures. In medical imaging, attribute misalignment could result in misdiagnosis by inaccurately describing clinical indicators. In assistive AI, especially for visually impaired users, relational or object misalignment may lead to incorrect environmental descriptions, potentially putting users at risk. These high-stakes scenarios underscore the urgent need to bridge the gap between benchmark-driven progress and deployment-focused validation. It is essential that future work moves beyond controlled datasets and begins testing and adapting misalignment mitigation methods in real-world settings to ensure their reliability, safety, and practical utility.

\end{document}